%% file: acl_latex.tex
\pdfoutput=1
\PassOptionsToPackage{table,dvipsnames}{xcolor}  

\documentclass[11pt,table,dvipsnames]{article}

\usepackage[final]{acl}
\usepackage{caption}
\usepackage{fontawesome}
\usepackage{times}
\usepackage{latexsym}
\usepackage{booktabs}
\usepackage{multirow}
\usepackage{algorithm}
\definecolor{headergray}{gray}{0.9}
\definecolor{hallublue}{HTML}{E0F7FA}
\newcommand{\rankA}[1]{\cellcolor{ForestGreen!70!white}{#1}} 
\newcommand{\rankB}[1]{\cellcolor{ForestGreen!55!white}{#1}} 
\newcommand{\rankC}[1]{\cellcolor{ForestGreen!40!white}{#1}} 
\newcommand{\rankD}[1]{\cellcolor{ForestGreen!25!white}{#1}} 
\newcommand{\rankE}[1]{\cellcolor{ForestGreen!15!white}{#1}} 

\usepackage{algpseudocode}

\usepackage{amsmath}
\usepackage{amssymb}

\usepackage[T1]{fontenc}
\usepackage{multicol}

\usepackage[T1]{fontenc}
\usepackage{bm}

\usepackage[utf8]{inputenc}
\usepackage{amsmath} 
\usepackage{microtype}

\usepackage{inconsolata}

\usepackage{pifont} 

\newcommand{\cmark}{\ding{51}} 
\newcommand{\xmark}{\ding{55}} 

\usepackage{graphicx}

\title{Teaching with Lies: Curriculum DPO on Synthetic Negatives for Hallucination Detection}

\author{
 \textbf{Shrey Pandit\thanks{Corresponding author: \href{mailto:shreypandit@utexas.edu}{shreypandit@utexas.edu}}\thanks{Equal contribution} },
 \textbf{Ashwin Vinod}\footnotemark[2] \\
 \textbf{Liu Leqi},
 \textbf{Ying Ding}
\\
\faGlobe~Webpage: \url{https://teachingwithlies.github.io/} \\
 The University of Texas at Austin
}

\begin{document}
\maketitle
\thispagestyle{plain}

\pagestyle{plain}
\pagenumbering{arabic}
\begin{abstract}
Aligning large language models (LLMs) to accurately detect hallucinations remains a significant challenge due to the sophisticated nature of hallucinated text. Recognizing that hallucinated samples typically exhibit higher deceptive quality than traditional negative samples, we use these carefully engineered hallucinations as negative examples in the DPO alignment procedure. Our method incorporates a curriculum learning strategy, gradually transitioning the training from easier samples, identified based on the greatest reduction in probability scores from independent fact checking models, to progressively harder ones. This structured difficulty scaling ensures stable and incremental learning. Experimental evaluation demonstrates that our HaluCheck models, trained with curriculum DPO approach and high quality negative samples, significantly improves model performance across various metrics, achieving improvements of upto 24\% on difficult benchmarks like MedHallu and HaluEval. Additionally, HaluCheck models demonstrate robustness in zero-shot settings, significantly outperforming larger state-of-the-art models across various benchmarks.

\end{abstract}

\section{Introduction}

\begin{figure}[ht]
\centering
\includegraphics[width=1\linewidth]{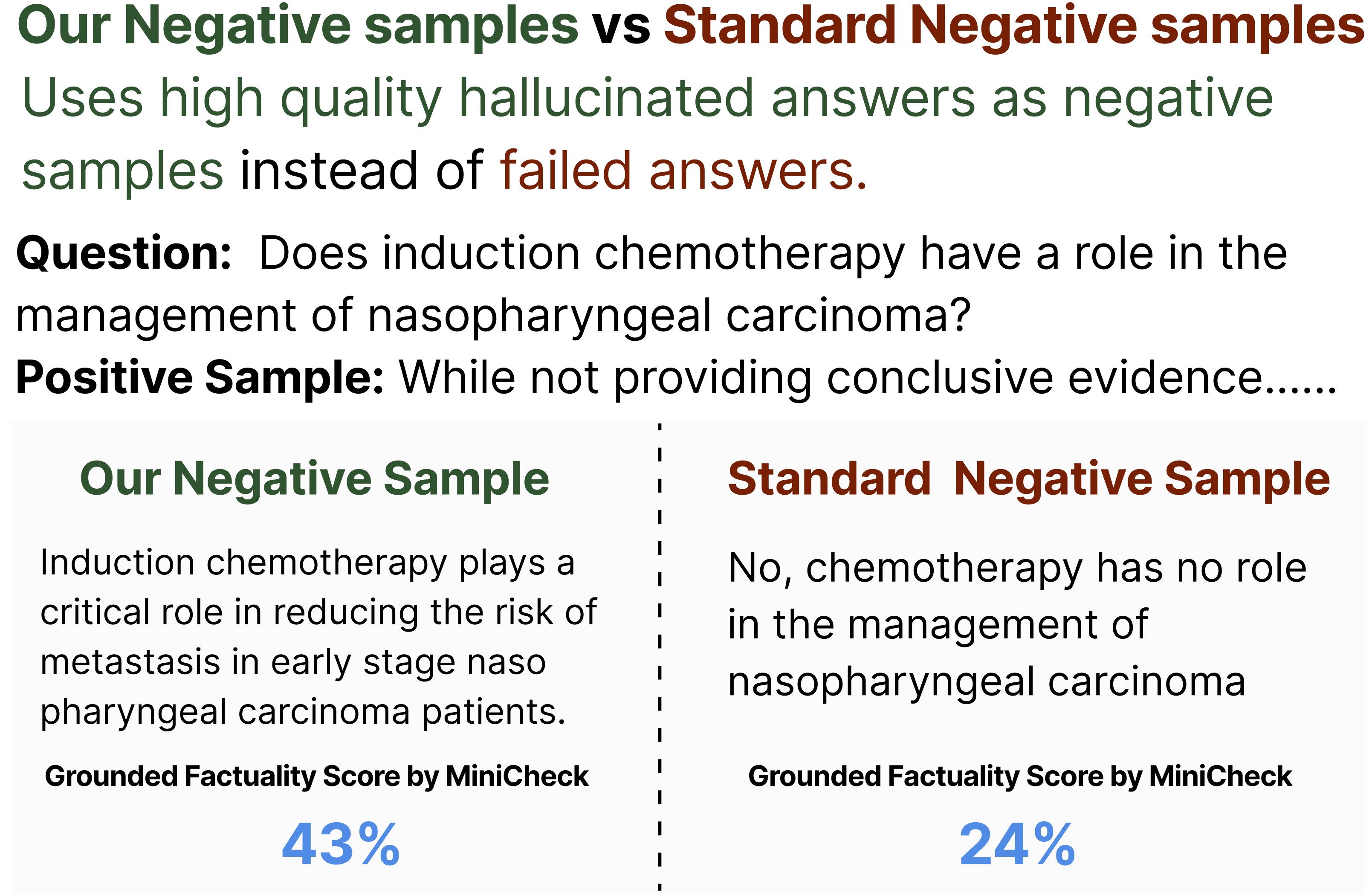}
\caption{Illustration of the qualitative difference between standard negative samples used in conventional DPO alignment and our proposed method, which leverages carefully curated hallucinated answers as high-quality negative examples in DPO alignment.}
\label{fig:Fig1}
\end{figure}

Large language models (LLMs) have achieved impressive performance across numerous NLP tasks, yet their deployment is limited by a tendency to produce fluent but factually incorrect “hallucinations.” Such errors erode trust and carry serious risks in domains with LLM applications like healthcare~\citep{singhal2022large}, software-development~\citep{softwarellm} and Law~\citep{llm_law}. Although various detection and mitigation strategies often based on external fact‐checkers or simplistic negative samples have been proposed, they struggle to identify sophisticated, plausibly crafted falsehoods.

\begin{figure*}[ht]
\centering
\includegraphics[width=1\linewidth]{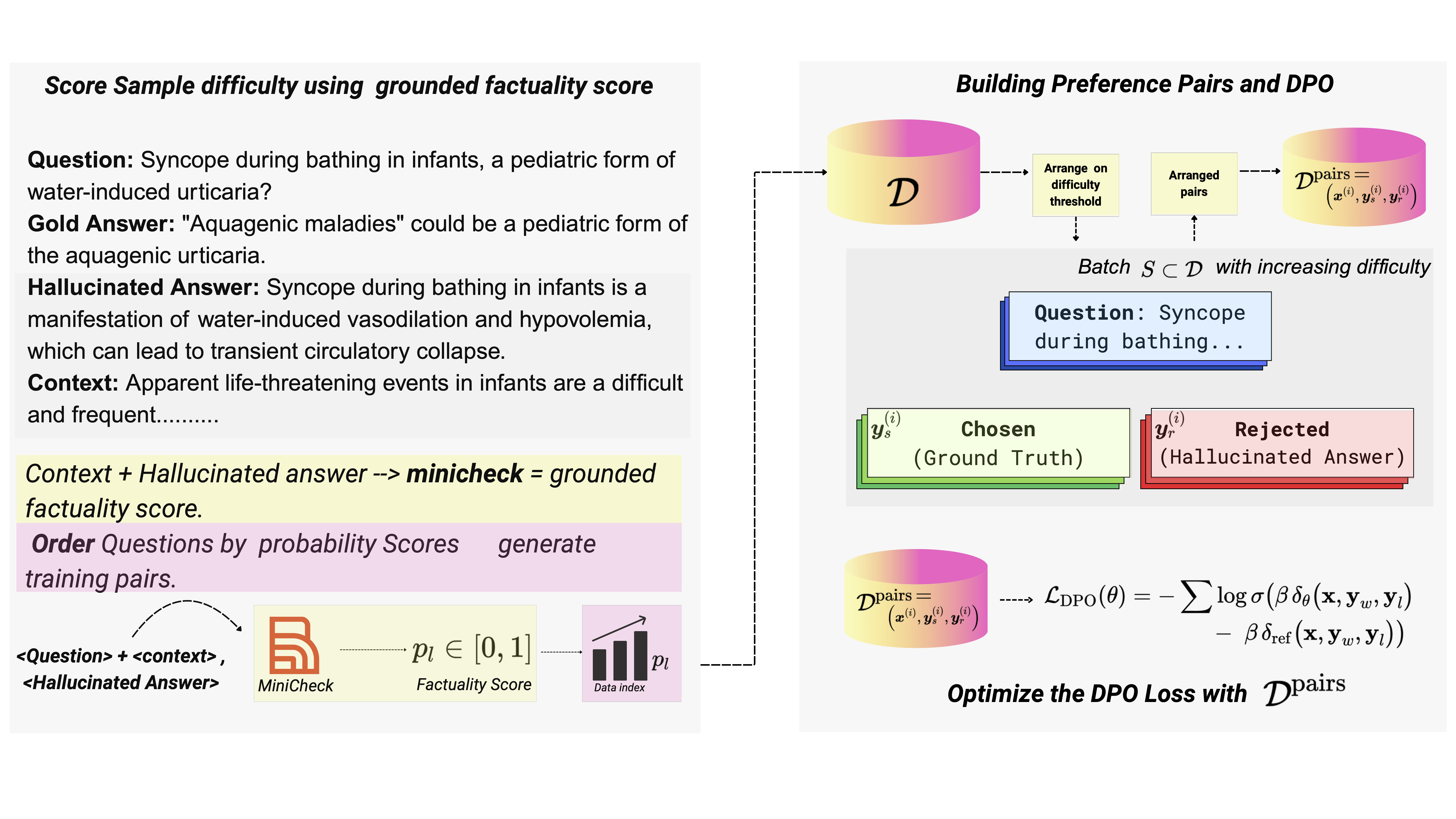}
\caption{Figure showing the pipeline for selecting high‑quality hallucinated negatives for Direct Preference Optimization (DPO). Each question and context is paired with a hallucinated answer and scored for grounded factuality via MiniCheck, then ranked by difficulty. In each batch, gold references (chosen) and top‑ranked hallucinations (rejected) form preference pairs. These pairs optimize the DPO objective, ensuring training against vetted, high‑quality negatives rather than arbitrary failures.}
\label{fig:pipeline}
\end{figure*}

To address these challenges, we introduce a novel alignment strategy leveraging Direct Preference Optimization (DPO)~\citep{dpo}, enhanced through a curriculum learning~\citep{cur1}~\citep{cur2} approach specifically tailored for hallucination detection. Our approach incorporates high quality hallucinated samples as negative samples into the alignment process instead of the usual low quality negative samples that are often selected from failed generations. 

We introduce \textbf{HaluCheck}, a family of Hallucination detection LLMs at two scales aligned via our curriculum-based DPO framework. We conduct extensive evaluations on the MedHallu~\citep{pandit2025medhallucomprehensivebenchmarkdetecting} and HaluEval~\citep{Hallueval} benchmarks and zero-shot evaluation on DROP, CovidQA, and PubMedQA, demonstrating that HaluCheck substantially outperforms existing baselines, including the widely adopted Llama-3.2 (1B and 3B) models. Notably, HaluCheck 3B yields up to a 24\% relative gain across core detection metrics (accuracy, precision, recall, and F1-score), while remaining competitive with far larger models such as GPT-4o. Our contributions are summarized as follows:

\begin{enumerate}
  \item We introduce a novel curriculum based sampling strategy that progressively selects hallucinated samples of increasing difficulty ranges obtained from fact verification models to enhance alignment training.
    \item We introduce \textbf{HaluCheck}, a suite of 1B–3B parameter models aligned with our DPO curriculum that leverages high-quality negative samples to deliver hallucination detection gains outperforming state of the art LLMs.
  \item Results demonstrate strong transferability of HaluCheck across multiple benchmarks and domains (Sec.~\ref{Results}), including zeroshot evaluation, confirming robustness in hallucinations detection task on diverse datasets.
\end{enumerate}

\input{latex/main_table_try}
\vspace{-3mm}

\section{Related Works}
\paragraph{Finetuning Models for Hallucination Detection} Recent research shows that both model-centric fine-tuning and sampling-based methods effectively detect hallucinations. LYNX~\citep{ravi2024lynx}, an open-source detector refined with distilled chain-of-thought reasoning, outperforms closed-source alternatives and provides HaluBench\cite{ravi2024lynx}, a diverse benchmark of semantically perturbed hallucinations. FACTCHECKMATE~\citep{alnuhait2024factcheckmate} preemptively flags hallucination risks via a lightweight MLP on hidden states and uses an intervention network to boost factuality with minimal overhead. SelfCheckGPT~\citep{manakul2023selfcheckgpt} requires no output probabilities or external knowledge: it samples multiple outputs and applies consistency measures such as BERTScore~\citep{zhang2019bertscore} at both sentence and passage levels. Existing work does not exploit alignment methods such as DPO~\citep{dpo}, despite their proven effectiveness. We introduce the first DPO approach that leverages curated hallucinated negatives, markedly improving hallucination detection.

\paragraph{Hallucination Detection Task} Hallucination in large language models (LLMs) has been extensively documented across various natural language processing tasks, such as machine translation~\citep{lee2019hallucinations_13RW}, dialogue systems~\citep{balakrishnan-etal-2019-constrained_14RW}, text summarization~\citep{durmus-etal-2020-feqa_15RW}, and question answering~\citep{sellam2020bleurtlearningrobustmetrics_16RW}, as detailed in recent survey literature~\citep{Ji_2023_12RW}. Benchmarks like Hades~\citep{liu2022tokenlevelreferencefreehallucinationdetection} and HaluEval~\citep{Hallueval} offer strong hallucination-detection protocols, and MedHallu~\citep{pandit2025medhallucomprehensivebenchmarkdetecting} provides carefully crafted adversarial answers that are ideal for our alignment approach. For the purpose of this work we choose MedHallu and HaluEval for the DPO alignment, as they have high quality hallucinated samples. Our proposed method is agnostic of task, and can be extended to other hallucination detection tasks like in summarization and dialogue answering setting.

\section{Hallucination Detection and Alignment} 
\paragraph{Problem formulation} For each sample \(i\) we define
Let \(\bm{x}^{(i)}\) denote the detection prompt (context + question + task instruction), \(\bm{y}^{(i)}_{\text{hall}}\) represent the \emph{hallucinated} class completion, and \(\bm{y}^{(i)}_{\text{true}}\) represent the \emph{factual} class completion. We define \(l^{(i)} \in \{0,1\}\) as the gold label, where a value of 1 indicates hallucination. From every labelled example we obtain a \textbf{preference pair} $(\bm{x}^{(i)},\bm{y}^{(i)}_{w},\bm{y}^{(i)}_{l})$, where
\vspace{-3pt}
\[
\begin{aligned}
(\bm{y}^{(i)}_{w},\bm{y}^{(i)}_{l}) &=
\begin{cases}
\bigl(\bm{y}^{(i)}_{\text{true}},\bm{y}^{(i)}_{\text{hall}}\bigr)
\end{cases}
\end{aligned}
\]

\paragraph{MiniCheck-Based Grounding Difficulty scoring} Before curriculum partitioning, we evaluate how well each hallucinated output is supported by its context using MiniCheck~\citep{tang2024minicheck}. For each example \((\bm{x}^{(i)}, \bm{y}^{(i)}_{\mathrm{hall}})\), we treat \(\text{question}=\bm{y}^{(i)}_{\mathrm{hall}}\) and \(\text{context}=\bm{x}^{(i)}\), and compute the grounding probability
\[
p_l^{(i)} \;=\; \mathcal{F}\bigl(\,\text{question} = \bm{y}^{(i)}_{\mathrm{hall}}\;\bigm|\;\text{context} = \bm{x}^{(i)}\bigr).
\]
We then use \(p_l^{(i)}\) to score difficulty and drive our curriculum stages. After sorting all examples by $p_l^{(i)}$ \text{ (ascending),}
\(\{\mathcal{B}_s\}_{s=1}^{S} \leftarrow \text{split into } S \text{ bins}\).
Lower \(p_l\) indicates easier hallucination cases, ensuring the curriculum starts with easy (high-grounding) and gradually moves to harder ones.

\paragraph{DPO Objective for Hallucination Detection} Let \(\pi_{\theta}\) be the current policy and \(\pi_{\text{ref}}\) the frozen reference model.
With trust–region parameter \(\beta\), and \(\sigma(z)=1/(1+e^{-z})\) the batch loss is:
\begin{align}
\mathcal{L}_{\text{DPO}}(\theta)=
&-\!\!\sum_{(\bm{x},\bm{y}_{w},\bm{y}_{l})\in\mathcal{B}}
\log\sigma\!\Bigl(
   \beta\bigl[\log\pi_{\theta}(\bm{y}_{w}\!\mid\!\bm{x}) \nonumber \\
   &-\log\pi_{\theta}(\bm{y}_{l}\!\mid\!\bm{x})\bigr]
   -\beta\bigl[\log\pi_{\text{ref}}(\bm{y}_{w}\!\mid\!\bm{x}) \nonumber \\
   &-\log\pi_{\text{ref}}(\bm{y}_{l}\!\mid\!\bm{x})\bigr]
\Bigr).
\label{eq:dpo-loss}
\end{align}

\noindent We provide a detailed algorithm for this pipeline in the supplementary (Alg. \ref{algorithm})

\section{Experimental Setup}
We describe the setup in the following section, and have a detailed section in supplementary \ref{Detailed Experimental setup} and \ref{Implementation details}
\paragraph{Model \& Datasets} We fine-tune \texttt{Llama-3.2} backbones (1 B and 3 B parameters) with LoRA adapters under the Direct Preference Optimization objective, using a joint corpus drawn from MedHallu and HaluEval. Hallucination detection is cast as binary classification via task-specific prompts.

\paragraph{Sampling Strategy \& Curriculum Learning} Negative examples are high-quality hallucinations scored by the MiniCheck fact-verifier. We sort them by decreasing MiniCheck confidence drop and train with a curriculum that proceeds from the easiest to the hardest negatives, yielding smoother and more robust convergence.


\input{latex/zeroshot}

\section{Results} \label{Results}
In the upcoming sections, \ref{baseline} \ding{202} we demonstrate that our HaluCheck models (1B and 3B) significantly outperform foundation LLMs despite their smaller size. In Sec.\ref{Zeroshot}, we further show that \ding{203} HaluCheck generalizes effectively to unseen datasets in a zero-shot setting, clearly outperforming its baseline model. In Sec.\ref{standardneg}, we validate the importance of using curated hallucinated samples rather than standard failed generations as negatives in DPO, showing that \ding{204} our model trained with curated hallucinated answers as negatives achieves superior performance. Finally, in Sec.~\ref{individual} and \ref{randomdpo}, we conduct ablations demonstrating HaluCheck's superior transferable skills when trained on individual datasets, and highlight the benefits of curriculum-based sampling over random selection.
\subsection{HaluCheck vs Baseline} \label{baseline}
As presented in Table~\ref{tab:results} \textbf{HaluCheck 3B}, trained with DPO on hallucinated answers as high quality negative samples, significantly outperforms similar and larger sized models. On HaluEval, it achieves an F1-score of \texttt{0.753}, surpassing the baseline \texttt{LLama-3.2 3B} (F1: \texttt{0.726}). On MedHallu, it outperforms the base model by \texttt{+26\%} F1 gain. Similarly, \textbf{HaluCheck 1B} shows strong performance on MedHallu (F1: \texttt{0.711}), while baseline LLama-3.2 1B lags behind (F1: \texttt{0.366}).
\ding{202} These results highlight our curriculum-based DPO approach's efficacy in enhancing hallucination detection while maintaining computational efficiency.

\subsection{Zero-shot evaluation} \label{Zeroshot}
\vspace{-1mm}
To gauge out‑of‑domain robustness, we ran a strict zeroshot test of \textbf{HaluCheck 3B} without any extra tuning or prompt changes against the backbone model \texttt{Llama-3.2 3B} and much larger \texttt{GPT-3.5-Turbo} on three external QA style hallucination benchmarks taken from the HaluBench dataset~\citep{ravi2024lynx}: \textsc{DROP}~\citep{dua2019drop}, \textsc{CovidQA}~\citep{moller2020covidqa}, and \textsc{PubMedQA}~\citep{jin2019pubmedqa}. As shown in Table~\ref{tab:accuracy_comparison}, HaluCheck 3B outperforms the Llama 3.2 3B model across the board, improving accuracy by \textbf{+4.8\%}, \textbf{+6.4\%}, and \textbf{+2.5\%} on the respective datasets, and also outperforming the \texttt{GPT-3.5-Turbo} on CovidQA by a substantial margin. \ding{203} These consistent gains achieved affirm that our curriculum based DPO alignment with using hallucinated samples as a high quality negative samples confers transferable hallucination detection skills that scale to unseen datasets.

\input{latex/stats_stdneg}

\subsection{DPO using Hallucinated vs Standard negative samples} \label{standardneg}

We show the importance of choosing curated hallucinated answers as a negative sample for DPO alignment by comparing the performance of Llama-3.2 3B model trained with standard negative samples. We sample these standard negative samples, by querying LLM for the question, and keeping the failed answers as negative samples, that is generally chosen as negative samples for DPO. We report the results in Table \ref{tab:medhallu_sn_vs_halucheck}, which clearly indicates that \ding{204} HaluCheck outperforms the later trained model. Also, to further back this choice, we report the grounded factuality score for the hallucinated answers from MedHallu and the standard negative samples we created, in Table \ref{tab:factuality_negatives}, showing the superiority of the samples as negatives for DPO, thereby being a better choice for DPO.

\vspace{-1mm}
\section{Conclusion}
\vspace{-1mm}
We present \textbf{HaluCheck} a curriculum-guided Direct Preference Optimization (DPO) framework for training an LLM for reliable hallucination detection task. A key contribution lies in replacing generic, model-generated failures with carefully curated, difficulty-ranked hallucinated samples as negative preferences during DPO alignment. This structured curriculum yields consistent gains, outperforming larger state-of-the-art models on multiple benchmarks and zero-shot tasks. Ablation results further validate that difficulty-aware negative sampling markedly strengthens the robustness of smaller language models.

\section*{Limitations}
Our proposed approach, while effective, exhibits certain limitations worth acknowledging. The curriculum-based Direct Preference Optimization (DPO) heavily relies on the quality and accuracy of the external fact-verification model (MiniCheck), potentially propagating any inherent biases or inaccuracies into our training process. Furthermore, our evaluations primarily focus on hallucinations within question-answering contexts, leaving unexplored the effectiveness in other NLP tasks such as dialogue generation, summarization, or multilingual settings. Additionally, treating hallucination detection purely as a binary classification task restricts the model's ability to identify partial or span-level hallucinations, thus limiting fine-grained interpretability. Lastly, although zeroshot evaluations suggest good generalization, there remains a risk of overfitting to dataset-specific adversarial patterns used during training, which may affect broader applicability and robustness.

\section*{Ethics statement}

Our work develops \textbf{HaluCheck} to improve reliable detection of hallucinations in LLM outputs, with the goal of reducing the risk of disseminating misleading or harmful information.  Our work uses publicly available MedHallu, and HaluEval data under MIT licenses We acknowledge that our reliance on an external fact‐verification model may introduce its own biases, and users should avoid treating automated detectors as infallible; human oversight remains essential, especially in high-stakes domains like healthcare or law. We encourage ongoing evaluation for fairness and transparency, and recommend that practitioners combine our approach with diverse verification methods to mitigate unintended biases or misuse.

\bibliography{refs} 

\clearpage
\appendix
\input{latex/supp}

\end{document}

%% file: latex/main_table_try.tex
\begin{table*}[ht]
  \small
  \centering
  \begin{tabular}{@{}l c ccc ccc@{}}
    \toprule
    \multirow{2}{*}{\textbf{Model}}
      & \multirow{2}{*}{\textbf{Average F1}}
      & \multicolumn{3}{c}{\textbf{MedHallu}~\citep{pandit2025medhallucomprehensivebenchmarkdetecting}}
      & \multicolumn{3}{c}{\textbf{HaluEval}~\citep{Hallueval}} \\
    \cmidrule(lr){3-5} \cmidrule(lr){6-8}
      & 
      & F1       & Precision & Accuracy
      & F1       & Precision & Accuracy \\ 
    \midrule
    Qwen-2.5 1.5B        & 0.464 & 0.227 & 0.642 & 0.525 & \rankE{0.701} & 0.568 & 0.610 \\
    LLama-3.2 1B         & 0.237 & 0.108 & 0.406 & 0.494 & 0.366 & 0.450 & 0.466 \\
    Qwen-2.5 3B          & \rankD{0.638} & \rankE{0.606} & 0.495 & 0.492 & 0.671 & 0.506 & 0.512 \\
    LLama-3.2 3B         & 0.612 & 0.499 & \rankD{0.696} & \rankE{0.566} & \rankD{0.726} & \rankE{0.743} & \rankD{0.732} \\
    LLama-3.1 8B         & 0.571 & 0.522 & \underline{\rankB{0.791}} & \rankD{0.608} & 0.620 & \textbf{\rankA{0.903}} & \rankE{0.711} \\
    Qwen-2.5 14B         & \rankC{0.720} & \rankD{0.619} & \rankE{0.691} & \rankC{0.633} & \rankB{\underline{0.821}} & \rankC{\underline{0.862}} & \rankB{\underline{0.829}} \\
    GPT 4o               & \textbf{\rankA{0.799}} & \rankB{\underline{0.737}} & \rankC{0.723} & \rankB{\underline{0.772}} & \textbf{\rankA{0.862}} & \rankB{\textbf{0.896}} & \rankA{\textbf{0.867}} \\
    \midrule
    HalluCheck-Llama 1B  & \rankE{0.637} & \rankC{0.664} & 0.511 & 0.527 & 0.611 & 0.481 & 0.468 \\
    HalluCheck-Llama 3B  & \rankB{\underline{0.756}} & \rankA{\textbf{0.759}} & \rankA{\textbf{0.845}} & \rankA{\textbf{0.782}} & \rankC{0.753} & \rankD{0.857} & \rankC{0.767} \\
    \bottomrule
  \end{tabular}
  \caption{Performance comparison of various models on the MedHallu and HaluEval hallucination detection benchmarks. Our proposed HaluCheck variants (1B and 3B) consistently outperform significantly larger foundational models. Notably, HaluCheck 3B demonstrates superior or comparable performance across both benchmarks, highlighting its efficiency and effectiveness despite its smaller size. Best scores are \textbf{bold}, runners-up are \underline{underlined.}}
  \label{tab:results}
\end{table*}

%% file: latex/zeroshot.tex
\begin{table}[t]
\centering
\small
\begin{tabular}{lcccc}
\toprule
\textbf{Model}        & \textbf{DROP} & \textbf{CovidQA} & \textbf{PQA} & \textbf{Avg} \\
\midrule
Llama 3.2 3B          & 52.50 & 56.10 & 55.20 & 54.60 \\
HaluCheck 3B          & \textbf{57.30} & \textbf{62.50} & 57.70  & \textbf{59.16} \\
GPT-3.5-Turbo         & 57.20 & 56.70 & \textbf{62.80} & 58.90 \\
\bottomrule
\end{tabular}
\caption{Accuracy (\%) on DROP, CovidQA and PQA (PubMedQA) for the baseline Llama 3.2 3B, our HaluCheck 3B, and GPT-3.5-Turbo (results from HaluBench~\citep{ravi2024lynx}). Results indicate strong performance of HaluCheck in zeroshot setting.} 
\label{tab:accuracy_comparison}
\end{table}

%% file: latex/stats_stdneg.tex
\begin{table}[t]
\centering
\scriptsize
\setlength{\tabcolsep}{4pt}
\begin{tabular}{l cc cc cc}
\toprule
\multirow{2}{*}{\textbf{Sample Type}} &
\multicolumn{2}{c}{\textbf{Easy}} &
\multicolumn{2}{c}{\textbf{Medium}} &
\multicolumn{2}{c}{\textbf{Hard}} \\
\cmidrule(lr){2-3}\cmidrule(lr){4-5}\cmidrule(lr){6-7}
& Mean & Median & Mean & Median & Mean & Median \\
\midrule
Standard Negative        & 0.282 & 0.202 & 0.273 & 0.201 & 0.248 & 0.182 \\
Our Hallucinated     & 0.303 & 0.202 & 0.379 & 0.269 & 0.391 & 0.294 \\
\bottomrule
\end{tabular}
\caption{Grounded factuality scores (MiniCheck \texttt{true\_prob}; higher is harder to spot) for standard negatives versus our curated hallucinated negatives, averaged over difficulty tiers for MedHallu dataset. The curated set provides consistently higher means and medians, confirming its superiority as training negatives for DPO.}
\label{tab:factuality_negatives}
\end{table}

%% file: latex/supp.tex
\input{latex/ablation_dataset}

\section{Ablations}
\subsection{Training on individual datasets} \label{individual}
\paragraph{Only Train on \texttt{MedHallu}}  

When we fine-tune the \texttt{HaluCheck-Llama-3B} detector exclusively on the \texttt{MedHallu} DPO set, the model achieves strong in-domain performance, with an \texttt{F1} of \texttt{0.729}, precision of \texttt{0.892}, and accuracy of \texttt{0.784} on the \texttt{MedHallu} benchmark. However, this specialization comes at the expense of generalization: when evaluated on \texttt{HaluEval}, the same model’s \texttt{F1} drops to \texttt{0.627}, precision to \texttt{0.578}, and accuracy to \texttt{0.593}. These results demonstrate that training solely on one dataset leads to overfitting to its particular style and content, limiting cross-dataset transfer.

\paragraph{Only Train on \texttt{HaluEval}}  
Conversely, training exclusively on the \texttt{HaluEval} DPO set yields a model that excels on \texttt{HaluEval} (\texttt{F1} = \texttt{0.793}, precision = \texttt{0.794}, accuracy = \texttt{0.793}), but underperforms on \texttt{MedHallu} (\texttt{F1} = \texttt{0.675}, precision = \texttt{0.623}, accuracy = \texttt{0.644}). Although the in-domain metrics on \texttt{HaluEval} are highest among the single-dataset trainings, the drop in \texttt{MedHallu} performance again highlights the narrow adaptation of the model to the peculiarities of its training set.

Training on each dataset in isolation yields high in-domain accuracy but poor transfer. In contrast, combining both DPO sets produces a model that maintains strong performance across \texttt{MedHallu} and \texttt{HaluEval}, underscoring the importance of diverse hallucination examples for robust detector alignment.  

\begin{figure}[h]
\centering
\includegraphics[width=1\linewidth]{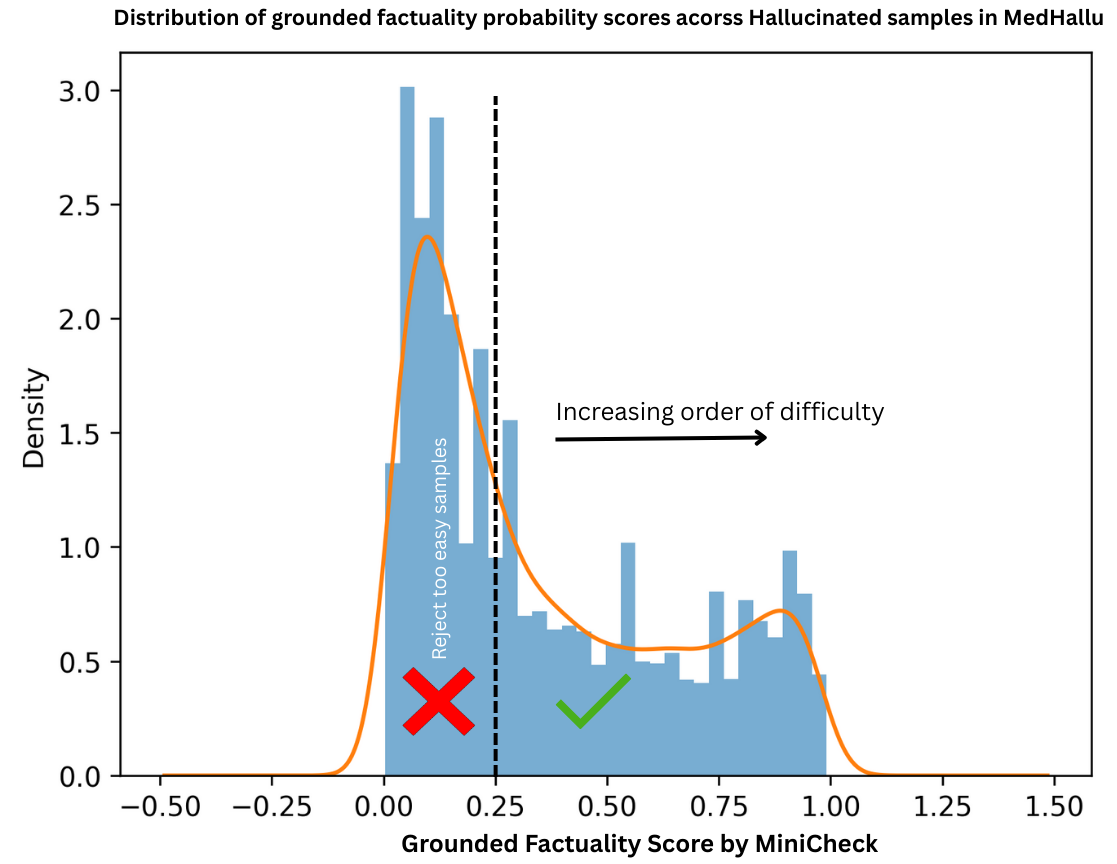}
\caption{Figure showing the grounded factuality of the hallucinated samples from MedHallu dataset. We keep only the samples that have a score above 0.25.}
\label{fig:difficulty}
\end{figure}

\input{latex/algorithm1}

\subsection{Random vs Curriculum learning DPO} \label{randomdpo}

\input{latex/curr_rand}

As Table~\ref{tab:curriculum_vs_random_f1} shows, replacing the usual \textit{random} selection of negative samples with a \textit{curriculum} that feeds the model increasingly difficult hallucinations produces a clear performance boost on both benchmarks and at both parameter scales. With just 1 B parameters, curriculum guided DPO lifts F1 on MedHallu 0.528 for the random baseline to 0.664 and on HaluEval from 0.446 to 0.611 gains that transform a lightweight detector from marginal to competitive accuracy. The effect is even more pronounced at 3 B curriculum training drives MedHallu F1 to 0.759 and HaluEval F1 to 0.753, surpassing the random counterpart by a wide margin and closing much of the gap to models an order of magnitude larger. These results confirm the intuition that hard, well vetted negatives presented in a staged fashion teach the model subtler decision boundaries than a grab-bag of arbitrary failures, leading to more robust hallucination detection with no increase in parameter count or compute budget.

\section{Additional Related Works}
\paragraph{Curriculum learning} 
Curriculum learning represents a training paradigm that strategically presents data samples in a meaningful sequence, effectively managing and optimizing the information a model encounters at each training step \cite{elman1993learning,bengio2009curriculum}.
Research has demonstrated the effectiveness of progressing from simple to complex examples across various NLP tasks, including language modeling \cite{choudhury2017curriculum,xu2020curriculum}, reading comprehension \cite{tay2019simple}, question answering \cite{sachan2016easy}, and machine translation \cite{zhang2019curriculum}.
In the context of LLM alignment, curriculum learning applications remain limited, with \cite{pattnaik2024curry} applying curriculum learning principles within the DPO framework for alignment.
\section{Detailed experimental setup} \label{Detailed Experimental setup}
\subsection{Model and Dataset Details}
We adopt the publicly released \texttt{Llama-3.2} checkpoints at two scales (1 B and 3 B parameters).  LoRA hyper-parameters follow \citet{hu2022lora}: rank{=}8, $\alpha{=}32$, dropout{=}0.05, and target modules \texttt{q\_proj}, \texttt{k\_proj}, \texttt{v\_proj}, and \texttt{o\_proj}.  
Training data comprise 9 000 examples from MedHallu’s \texttt{pqa\_artificial} split plus 8 000 items (80 \%) from the HaluEval training partition, forming 17 000 DPO preference pairs.  
Evaluation is conducted on the 1 000-example MedHallu \texttt{pqa\_labeled} set and the held-out 2 000 HaluEval test items.

\input{latex/split_results}

\subsection{Curriculum Construction}
For every hallucinated answer $h_i$ paired with context $c_i$, the MiniCheck verifier returns a grounding probability $p_i$. Examples with $p_i < 0.25$ (very poor grounding) are discarded. 
The remainder are sorted by ascending values of $p_i$.
DPO training proceeds batch wise on the sorted data for four epochs, with all batches trained per epoch, thereby gradually exposing the model to increasingly difficult negatives.  
Table~\ref{tab:cutoff_ablation_full} in the main paper reports ablations over alternative cut-offs; the chosen $0.25$–$1.0$ range yields the highest F1 scores, consistent with the grounded factuality distribution visualized in Figure~\ref{fig:difficulty}.

\input{latex/StandNeg_table}

\section{Implementation details} \label{Implementation details}
Training was performed using Direct Preference Optimization (DPO) with hyperparameters set as follows: learning rate = $1 \times 10^{-5}$, beta = 0.1, gradient accumulation steps = 4, per-device batch size = 4, and total epochs = 25. We used a paged AdamW optimizer with 8-bit quantization and mixed-precision training (FP16) for computational efficiency. Sequential sampling was used during training to maintain curriculum learning order. The model’s performance was periodically assessed on the MedHallu labeled validation set. Evaluation metrics included accuracy, precision, recall, and F1-score, computed both overall and separately by difficulty (easy, medium, hard).

\section{LLMs Used in Discriminative Tasks}

\textbf{GPT-4o and GPT-4o mini.} GPT-4o~\cite{openai2024gpt4ocard} are a series of commercial LLMs developed by OpenAI. Renowned for their state-of-the-art performance, these models have been extensively utilized in tasks such as medical hallucination detection. Our study employs the official API provided by the OpenAI platform to access these models. For all other models below, we implement them through Hugging Face package.

\textbf{Llama-3.1 and Llama-3.2.} Llama-3.1 and Llama-3.2~\citep{grattafiori2024llama3herdmodels} are part of Meta's open-source multilingual LLMs, Llama 3.1 (July 2024) includes 8B, 70B, and 405B parameter models optimized for multilingual dialogue. Llama 3.2 (September 2024) offers 1B, 3B, 11B, and 90B models with enhanced accuracy and speed. We use Llama 3.2 1B and 3B models as our backbone for training DPO, and also use the Llama 3.1 8B model in our evaluation table for performance comparison

\textbf{Qwen2.5.} Qwen2.5~\citep{qwen2.5} is an advanced LLM designed to handle complex language tasks efficiently. It has been applied in various domains, including medical hallucination detection. We use the 3B, 7B and 14B variants in our work.

\section{Hardware Resources and Computational Costs}

During the DPO training process using LoRA, we primarily used the \texttt{Llama-3.2 1B} and \texttt{Llama-3.2 3B} model as a base model for our HaluCheck Model, running it for 12 hours on an NVIDIA RTX A6000 GPU with 48,685 MiB of RAM. Additionally, we employed models such as \texttt{Qwen2.5-1.5B, 3B, 14B}, and GPT models as evaluators for benchmarkings. To enhance the efficiency and speed of our code execution, we utilized software tools like \texttt{vLLM} and implemented batching strategies. These optimizations were critical for managing the computational load and ensuring timely processing of our experiments.


%% file: latex/ablation_dataset.tex
\begin{table*}[ht]
  \centering
  \small
  \begin{tabular}{@{}l cc ccc ccc@{}}
    \toprule
    \multirow{2}{*}{\textbf{Model}}
      & \multicolumn{2}{c}{\textbf{DPO set}}
      & \multicolumn{3}{c}{\textbf{MedHallu}}
      & \multicolumn{3}{c}{\textbf{HaluEval}} \\
    \cmidrule(lr){2-3} \cmidrule(lr){4-6} \cmidrule(lr){7-9}
      & MedHallu & HaluEval
      & F1 & Precision & Accuracy
      & F1 & Precision & Accuracy \\ 
    \midrule
    HalluCheck-Llama 3B
      & \cmark  & \xmark  
      & 0.729 & 0.892 & 0.784  
      & 0.627     & 0.578     & 0.593      \\
    HalluCheck-Llama 3B   
      & \xmark  & \cmark  
      & 0.675     & 0.623     & 0.644      
      & 0.793 & 0.794 & 0.793  \\
    HalluCheck-Llama 3B                       
      & \cmark  & \cmark  
      &   0.759     &   0.845     &    0.782    
      &    0.733    &    0.857    &    0.767     \\
    \bottomrule
  \end{tabular}
  \caption{Performance over training with different train sets.}
  \label{tab:table2}
\end{table*}

%% file: latex/algorithm1.tex
\begin{algorithm}[t]
\small
\caption{Curriculum-Based DPO Alignment for Hallucination Detection}
\begin{algorithmic}[1]
  \Require 
    Detection data $\{(\bm{x}^{(i)}, \bm{y}^{(i)}_{\text{true}}, \bm{y}^{(i)}_{\text{hall}}, l^{(i)})\}_{i=1}^N$, 
    fact-checker $\mathcal{F}$ (using MiniCheck, returns probability), 
    policy $\pi_{\theta}$, 
    frozen ref. policy $\pi_{\text{ref}}$, 
    stages $S$
  \Ensure 
    Fine-tuned detector $\pi_{\theta}$
  \medskip
  \State \textbf{\# Score difficulty}
  \For{each $(\bm{x}, \bm{y}_{\text{true}}, \bm{y}_{\text{hall}}, l)$}
    \State $p_l \gets \mathcal{F}(\bm{y}_l \mid \bm{x})$
  \EndFor
  \State \textbf{\# Partition into stages}
  \State sort by $p_l$ (asc.) and split into $\{\mathcal{B}_s\}_{s=1}^S$
  \State \textbf{\# Generate preference pairs}
  \For{$i=1,\dots,N$}
    \State $\bm{y}^{(i)}_{w} \gets \bm{y}^{(i)}_{\text{true}}$
    \State $\bm{y}^{(i)}_{l} \gets \bm{y}^{(i)}_{\text{hall}}$
    \State store $(\bm{x}^{(i)}, \bm{y}^{(i)}_{w}, \bm{y}^{(i)}_{l})$
  \EndFor
  \State \textbf{\# Stage-wise DPO fine-tuning}
  \For{$s=1,\dots,S$}
    \State Define:
    \State $\delta_\theta(\bm{x}, \bm{y}_{w}, \bm{y}_{l}) = 
            \log\pi_\theta(\bm{y}_{w} \mid \bm{x}) - 
            \log\pi_\theta(\bm{y}_{l} \mid \bm{x})$
    \State $\delta_{\text{ref}}(\bm{x}, \bm{y}_{w}, \bm{y}_{l}) = 
            \log\pi_{\text{ref}}(\bm{y}_{w} \mid \bm{x}) - 
            \log\pi_{\text{ref}}(\bm{y}_{l} \mid \bm{x})$
    \State Minimize over $(\bm{x}, \bm{y}_{w}, \bm{y}_{l}) \in \mathcal{B}_s$:
\Statex\hspace{\algorithmicindent}%
      \(\displaystyle
        \mathcal{L}_{\mathrm{DPO}}(\theta)
        = 
        \begin{aligned}[t]
          -\sum \log\sigma\bigl(&\,\beta\,\delta_\theta(\bm{x},\bm{y}_w,\bm{y}_l)\\
                                   &\;-\;\beta\,\delta_{\mathrm{ref}}(\bm{x},\bm{y}_w,\bm{y}_l)\bigr)
        \end{aligned}
      \)
  \EndFor
  \State \Return $\pi_{\theta}$
\end{algorithmic}
\label{algorithm}
\end{algorithm}

%% file: latex/curr_rand.tex
\begin{table}[ht]
\centering
\small
\begin{tabular}{l cc}
\toprule
\textbf{Model} & \textbf{MedHallu F1} & \textbf{HaluEval F1} \\
\midrule
HalluCheck 1B (Random) & 52.80     & 44.60    \\
HalluCheck 1B (Curr.)  & 66.40 & 61.10 \\

HalluCheck 3B (Random) & 69.40     & 63.10     \\
HalluCheck 3B (Curr.)  & 75.90 & 75.30 \\
\bottomrule
\end{tabular}
\caption{F1 comparison of curriculum-guided vs.\ random sampling for HalluCheck models on MedHallu and HaluEval.}
\label{tab:curriculum_vs_random_f1}
\end{table}

%% file: latex/split_results.tex
\begin{table*}[ht]
\small
\centering
\setlength{\tabcolsep}{4pt}  
\begin{tabular}{ll c ccc ccc}
\toprule
\multirow{2}{*}{\textbf{Split Range}} & \multirow{2}{*}{\textbf{Model}} & \multirow{2}{*}{\textbf{Avg F1}} &
\multicolumn{3}{c}{\textbf{MedHallu}} &
\multicolumn{3}{c}{\textbf{HaluEval}} \\
\cmidrule(lr){4-6}\cmidrule(lr){7-9}
 &  &  & F1 & Prec  & Acc 
 & F1  & Prec  & Acc  \\
\midrule
\multirow{2}{*}{$0.00$--$0.75$}      & HaluCheck 1B & 0.499 & 0.404 & 0.717 & 0.596 & 0.595 & 0.491 & 0.458 \\
                                  & HaluCheck 3B & 0.714 & 0.729 & 0.892 & 0.784 & 0.699 & 0.812 & 0.728 \\
\addlinespace
\multirow{2}{*}{$0.25$--$1.00$}      & HaluCheck 1B & 0.637 & 0.664 & 0.511 & 0.527 & 0.611 & 0.481 & 0.468 \\
                                  & HaluCheck 3B & 0.756 & 0.759 & 0.845 & 0.782 & 0.753 & 0.857 & 0.767 \\
\addlinespace
\multirow{2}{*}{$0.25$--$0.75$}   & HaluCheck 1B & 0.625 & 0.651 & 0.501 & 0.511 & 0.599 & 0.512 & 0.469 \\
                                  & HaluCheck 3B & 0.712 & 0.696 & 0.727 & 0.704 & 0.728 & 0.824 & 0.739 \\
\addlinespace
\multirow{2}{*}{$0.00$--$1.00$}         & HaluCheck 1B & 0.614 & 0.622 & 0.601 & 0.459 & 0.606 & 0.494 & 0.455 \\
                                  & HaluCheck 3B & 0.743 & 0.743 & 0.905 & 0.770 & 0.744 & 0.829 & 0.759 \\
\bottomrule
\end{tabular}
\caption{\textbf{Ablation over curriculum difficulty cut-offs}.  
Each split indicates the MiniCheck grounding-probability interval used when selecting hallucinated negatives. “Avg F1” is the mean F1 score across MedHallu and HaluEval; higher is better for all metrics.}
\label{tab:cutoff_ablation_full}
\end{table*}

%% file: latex/StandNeg_table.tex
\begin{table}[ht]
\centering
\small
\begin{tabular}{lccc}
\toprule
\textbf{Model} & \textbf{F1} & \textbf{Precision} & \textbf{Accuracy} \\
\midrule
HaluCheck 1B          & 0.664 & 0.511 & 0.527 \\
Llama-3.2 1B-SN       & 0.622 & 0.494 & 0.491 \\
\addlinespace
HaluCheck 3B          & 0.729 & 0.845 & 0.782 \\
Llama-3.2 3B-SN       & 0.691 & 0.772 & 0.717 \\
\bottomrule
\end{tabular}
\caption{\textbf{Hallucination detection on the MedHallu dataset}.  
“SN” models were aligned with standard negative samples in DPO, while HaluCheck models were aligned with curated hallucinated negatives. Higher is better on all metrics.}
\label{tab:medhallu_sn_vs_halucheck}
\end{table}